\documentclass{article}
\usepackage[preprint]{spconf}
\usepackage{amsmath,graphicx,array,tabularx}
\usepackage{booktabs}
\usepackage{xcolor}
\usepackage{listings}

\title{G-Augment: Searching for the Meta-Structure\\
of Data Augmentation Policies for ASR}

\name{Gary Wang, Ekin D. Cubuk, Andrew Rosenberg,}
\nameplus{Shuyang Cheng{$^{\dagger*}$}, Ron J. Weiss{$^{\ddagger*}$},}
\nameplusplus{ Bhuvana Ramabhadran, Pedro J. Moreno, Quoc V. Le and Daniel S. Park
\thanks{*Work done while at Google.}}
\address{Google Inc.,~~$\dagger$Pony.ai,~~$\ddagger$Meta Inc.}
\copyrightnotice{\footnotesize\begin{tabular}{p{0.965\textwidth}}
\\
\copyright\ Copyright 2023 IEEE. Published in the 2022 IEEE Spoken Language Technology Workshop (SLT) (SLT 2022), scheduled for 19-22 January 2023 in Doha, Qatar. Personal use of this material is permitted. However, permission to reprint/republish this material for advertising or promotional purposes or for creating new collective works for resale or redistribution to servers or lists, or to reuse any copyrighted component of this work in other works, must be obtained from the IEEE. Contact: Manager, Copyrights and Permissions / IEEE Service Center / 445 Hoes Lane / P.O. Box 1331 / Piscataway, NJ 08855-1331, USA. Telephone: + Intl. 908-562-3966.\end{tabular}}

\begin{document}

\maketitle

\begin{abstract}
Data augmentation is a ubiquitous technique used to provide robustness to automatic speech recognition (ASR) training. However, even as so much of the ASR training process has become automated and more ``end-to-end," the data augmentation policy (what augmentation functions to use, and how to apply them) remains hand-crafted. We present G(raph)-Augment, a technique to define the augmentation space as directed acyclic graphs (DAGs) and search over this space to optimize the augmentation policy itself. We show that given the same computational budget, policies produced by G-Augment are able to perform better than SpecAugment policies obtained by random search on fine-tuning tasks on CHiME-6 and AMI. G-Augment is also able to establish a new state-of-the-art ASR performance on the CHiME-6 evaluation set (30.7\% WER). We  further demonstrate that G-Augment policies show better transfer properties across warm-start to cold-start training and model size compared to random-searched SpecAugment policies.
\end{abstract}
\section{Introduction}
\label{sec:intro}

Data augmentation \cite{lecun1998gradient,simard2003best,shorten2019survey} is an important component of deep learning and has demonstrated to be a crucial component of training deep networks on a wide range of tasks, including automatic speech recognition (ASR) \cite{1169544mtr,deng2000large,ko2017study,ko2015audio,cui2015data,cui2015multilingual,jaitly2013vocal,park2019specaugment}.

While methods for automatically optimizing augmentation policies have been introduced and studied \cite{Ragni2014DataAF,cubuk2018autoaugment,cubuk2019randaugment}, previous studies made certain structural assumptions about how data augmentations are applied. For example, augmentation searches for images typically assume a hierarchy, where certain augmentations are assumed to always be applied in addition to other augmentation operations. The same has been true for ASR, where assumptions about the meta-structure of the augmentation are made before searching over parameters of the augmentations themselves \cite{wang2020scada}.

While such accumulated heuristics are effective for addressing tasks that have been studied extensively before with a set of well-known augmentations, when encountered with a new task or with a new set of augmentations, one needs to re-establish the heuristics for designing a good augmentation scheme. For example, in \cite{park2020specaugment}, the authors discovered that SpecAugment \cite{park2019specaugment} did not compose well with multi-style training augmentation \cite{1169544mtr,mtr}, and found that they needed to ensemble the augmentations to benefit from both.

In this work, we address this problem by a scheme we refer to as G(raph)-Augment, where a stochastic augmentation policy is parameterized by a directed acyclic graph (DAG) whose edges are labeled by sampling probabilities and augmentation parameters. By simultaneously searching for the graph structure and the parameters that label the graph, we are able to optimize not only the augmentation parameters of the individual augmentations, but how those augmentations are being applied. We utilize 17 ASR augmentations in our search space, details of which can be found in section \ref{sec:aug_list}. We use an evolutionary algorithm \cite{golberg1989genetic,holland1992adaptation} to optimize these graphs based on the dev-set performance of the augmentation.

The search is conducted on two ``warm-start" tasks, where we pre-train a Conformer \cite{conformer} RNN-T \cite{graves2012sequence} model on the SpeechStew \cite{speechstew} dataset and fine-tune on the CHiME-6 \cite{watanabe2020chime} and AMI \cite{ami} corpora. For the AMI task, we remove the AMI portion of the SpeechStew dataset for pre-training. We compare the performance of the best discovered G-Augment policy against the best SpecAugment policy found using random search with the same computational budget.%
\footnote{Naively, one may deem the comparison between a random search and a genetic algorithm to be unfair. We, however, must note that the search space size of G-Augment is much larger than that of SpecAugment (by a factor of $\gg 10^{50}$) in this work, which justifies the comparison in our view.}
By doing so, we are able to arrive at the following results:
\begin{itemize}
  \item The best G-Augment policy discovered outperforms the best SpecAugment policy on both tasks.
  \item The G-Augment policies exhibit better transfer properties across warm to cold-start training and model size than the SpecAugment policies.
  \item By adapting the G-Augment policy for training a very large (1B parameter) Conformer \cite{conformer} pre-trained \cite{baevski2020wav2vec,park2020nst1,bigssl} on YouTube and SpeechStew, we achieve state-of-the-art performance on CHiME-6.
\end{itemize}

While we have limited our scope to ASR in this work, G-Augment is a general framework that can be applied to any task where augmentation is utilized.

\section{Related Works}
\label{sec:format}

Data augmentation is an effective method for improving the generalization performance of deep learning models. Domain specific augmentations have been utilized for a variety of domains, from image classification~\cite{simard2003best} to speech recognition~\cite{park2019specaugment}. More recently, automated data augmentation methods have been utilized for increasing the efficacy of data augmentation for 2D image classification, where a policy is (meta)-learned using a large search space of possible data augmentations~\cite{cubuk2018autoaugment,lemley2017smart,ratner2017learning}. Automated data augmentation methods have been made more efficient for image classification~\cite{cubuk2019randaugment,ho2019population,lim2019fast} and have been extended to other modalities in vision~\cite{zoph2019learning,cheng2020improving}. Some efforts to apply automated augmentation search for ASR tasks has also shown success \cite{wang2020scada,hu2021sapaugment}.

While these previous attempts learned augmentation parameters such as application probability and distortion magnitude from the data, they used a manually chosen augmentation structure. For example, AutoAugment learned 25 subpolicies, each with two layers of augmentation. Decisions such as whether flips and cropping should be applied were manually made specific to the dataset~\cite{cubuk2019randaugment,ho2019population,lim2019fast}. For example, Cutout~\cite{devries2017improved} was always applied after the AutoAugment layers on CIFAR-10~\cite{krizhevsky2009learning}, but not on reduced SVHN~\cite{netzer2011reading}. In this work, we fully automate the optimization of the data augmentation policy: the graph structure of the policy and parameters of individual augmentations that make up the policy are learned jointly. In addition, we are able to demonstrate positive transfer of augmentation policies from the search task to set-ups with different training dynamics and model sizes. To our knowledge, our work is the first example of an automated data augmentation approach that outperforms manually designed policies in speech recognition such as SpecAugment.

Methods for searching over graph spaces have been extensively investigated in the context of neural architecture search \cite{zoph2018learning, real2019regularized}. While we choose to use a relatively simple evolutionary algorithm \cite{golberg1989genetic,holland1992adaptation} to search over augmentation policies in this particular work, a variety of methods have been employed for such searches in the literature \cite{zoph2018learning,real2019regularized,koza1994genetic,stanley2002evolving,stanley2007compositional}, an extensive list of which can be found in \cite{dong2021automated}.

\section{G-Augment}
\label{sec:g-augment}

To search for both how the augmentations are applied and the parameters of the augmentations themselves, we parameterize an augmentation policy as a graph with labeled nodes and edges. Here we describe the details of this parameterization and the algorithm we employ to search over this space.

\subsection{Search Space}
\label{sec:g-augment:search_space}

\begin{figure*}
\begin{minipage}{0.56\linewidth}
    \includegraphics[width=\linewidth]{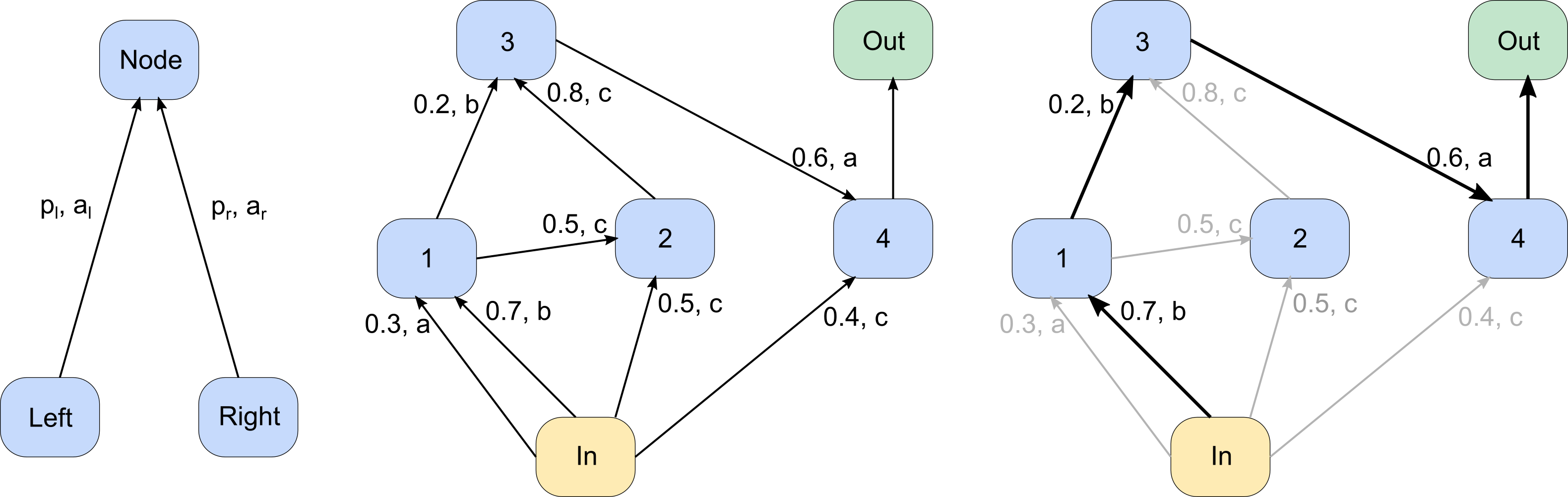}
\end{minipage}
\hfill
\begin{minipage}{0.41\linewidth}
\begin{lstlisting}[language=python,mathescape]
def node_state(node, batch):
  if node.is_input_node():
    return batch
  choice = tf.random.uniform(shape=batch.shape[0])
  choose_left = tf.cast(choice < node.p_left, tf.bool)  
  left_in = node.a_left(node_state(node.left_node, batch))
  right_in = node.a_right(node_state(node.right_node, batch))
  return tf.where(choose_left, left_in, right_in)

out_batch = node_state(LAST_ENSEMBLE_NODE, in_batch)
\end{lstlisting}
\label{fig:dag}
\end{minipage}
\caption{DAG structured stochastic augmentation policies (left) and pseudocode for applying an augmentation to an input batch using this structure (right). The first graph illustrates the notation used to express the relationship between nodes. A given ensemble node has two inputs, denoted the left and right nodes. The edges connecting these nodes are labeled by a selection probability and an augmentation. The second graph is an example of a policy graph, where three augmentations ($a$, $b$ and $c$) are utilized. Each path represents an augmentation operation whose sampling probability is obtained by the product of sampling probabilities of the edges of the path. For example, the bold path in the third graph represents the augmentation $\text{Out} = a(b(b(\text{In})))$ and is sampled with a probability of $0.7 \times 0.2 \times 0.6$.}
\label{f:layers}
\end{figure*}

We parameterize an augmentation by a directed acyclic graph (DAG), consisting of a single input node, a single output node and a number $N$ of ensemble nodes. The input node has only outgoing edges that can connect to ensemble nodes. The output node connects to a single ensemble node via an edge, which passes its state to the output. Each node represents an augmented state of the data, while the edges represent the augmentations themselves.

Each ensemble node of the graph takes two inputs (Figure \ref{fig:dag}). We denote one of the incoming edges the left edge and the other the right edge of a given ensemble node for convenience. We denote the node connected to the tail of the left/right edge as the left/right input, respectively. The incoming edges are labeled by sampling probabilities $p_l$ and $p_r$ that sum to unity, and quadruples $a_l$ and $a_r$ that represent augmentations. The state of a given node is obtained by applying the augmentations $a_l$/$a_r$ to the left/right inputs and sampling them with probability $p_l$/$p_r$ respectively. In other words,
\begin{equation}
\text{(node state)} =
\begin{cases}
a_l(\text{left input}) & \text{w/ probability $p_l$,} \\
a_r(\text{right input}) & \text{w/ probability $p_r$.}
\end{cases}
\end{equation}

We require the graph to be directed and all directed paths trace back to the input. This is enforced by assigning indices $1, \cdots, N$ to the ensemble nodes, the index 0 to the input node, and selecting the left/right input indices of node $n$ from $[0, n-1]$. The output node is always connected to node $N$.

The augmentation policy is applied to an input by stochastically back-propagating through the graph. Given an input, the augmentation to be applied to that input is determined by starting at the output node and back-propagating through the graph by randomly selecting the path to travel based on the selection probability of the edges. The path connecting the input node to the output node sampled this way represents an augmentation obtained by sequentially composing the augmentations encountered in the path. The probability of a particular path being selected is obtained by multiplying the selection probabilities. This process is depicted in figure \ref{fig:dag} along with pseudo-code. Any AutoAugment \cite{cubuk2018autoaugment} policy or RandAugment policy \cite{cubuk2019randaugment} is representable by such a graph.

To make the search space uniform, we represent an augmentation by a quadruple $a = (t, q, x_1, x_2)$ where $t$ is a string denoting the type of augmentation, $q$ is the application probability (not to be confused with the sampling probability) and $x_1$ and $x_2$ are the strength parameters for the augmentation. $x_i$ are taken to have 11 discrete integer values from 0 to 10. We employ 17 augmentations, which we list in section \ref{sec:aug_list}.

\subsection{Search Method}
\label{sec:g-augment:search_method}

We use a generational evolutionary algorithm \cite{golberg1989genetic,holland1992adaptation} with binary tournament selection \cite{miller1995genetic} to search the G-Augment search space, although other methods may be utilized \cite{koza1994genetic,stanley2002evolving,stanley2007compositional,zoph2018learning, real2019regularized}. We use the implementation by \cite{jaderberg2017population}.

A generational evolutionary algorithm is parameterized by a population size $P$ and a mutation rate $\mu$. For us, the population of the first generation is constructed by generating $P$ random policy graphs by uniformly sampling the search space. Given a population of generation $g$, the population for generation $g+1$ is constructed via a binary tournament. That is, $P$ pairs of population members are randomly drawn with replacement from the population of generation $g$ and the population member with better fitness within each pair is kept to survive. The population of generation $g+1$ is assembled by taking the surviving $P$ population members and applying mutation with mutation rate $\mu$. In our case, the policy that yields a lower dev WER on the search ASR task is deemed more ``fit" and survives the selection process.

The particular mutation we apply to graphs follows \cite{real2019regularized}. We first randomly select a single edge of the graph, randomly re-select its tail node within the allowed set of nodes, and finally reset its augmentation quadruple. We then mutate the selection probabilities and the $q$, $x_1$ and $x_2$ values of the augmentation parameters of all the other edges with mutation rate $\mu$. We define a mutation of an augmentation parameter $x_i$ to be a change of $\pm 1$ to its value and a mutation of a selection probability to be a change of $\pm 0.1$. For the application probabilities $q$, a mutation samples a random number uniformly in the range of $[-0.2, 0.2]$, adds it to the current probability and clips its value to the allowed range of $[0, 1]$.

\begin{table*}[h!]
\caption{Description of the 17 augmentation components utilized in our search space.}
\label{t:bblocks}
\vskip 0.1in
\centering
\resizebox{1.7\columnwidth}{!}{%
\begin{tabular}{cccccccccc}
\toprule
Code & Name & &
\multicolumn{3}{c}{$x_1$} & &
\multicolumn{3}{c}{$x_2$} \\
\cmidrule{3-5} \cmidrule{7-9}
& & &
parameter & range & scale & &
parameter & range & scale \\
\midrule
CO & Cut Out & &
mask size & [0, 30] & linear & &
density ratio & [0, 0.5] & linear
\\
FM & Frequency Mask & &
multiplicity & [0, 8] & linear & &
masking ratio & [0, 1.0] & linear
\\
FS & Frequency Shift & &
multiplicity & [0, 8] & linear & &
filter coverage & [0, 1.0] & linear
\\
FN & Frequency Noise& &
max stddev & [0, 0.5] & linear & &
- & - & -
\\
FW-L & Frequency Warp (Linear) & &
warp ratio & [0.0, 1.0] & linear & &
- & - & -
\\
FW-LG & Frequency Warp (Log) & &
warp ratio & [0.0125, 0.79] & log & &
- & - & -
\\
GN & Gaussian Noise & &
noise ratio & [0, 1.0] & linear & &
- & - & -
\\
Id & Identity & &
- & - & - & &
- & - & -
\\
RC & Random Convolution & &
filter freq size & [0, 50] & linear & &
filter time size & [0, 50] & linear
\\
TP & Time Perturb& &
max utterance ratio & [0, 0.6] & linear & &
- & - & -
\\
TM-AM & Time Mask (Adaptive Multiplicity)& &
multiplicity ratio & [0.001, 0.1] & log & &
- & - & -
\\
TM-AS & Time Mask (Adaptive Size)& &
size ratio & [0.001, 0.316] & log & &
- & - & -
\\
TM-FA & Time Mask (Fully Adaptive)& &
multiplicity ratio & [0.001, 0.1] & log & &
size ratio & [0.001, 0.316] & log
\\
TW-A & Time Warp (Adaptive)& &
utterance length ratio & [0.005, 0.5] & log & &
- & - & -
\\
TW & Time Warp (Non-Adaptive)& &
size ratio & [5, 500] & log & &
- & - & -
\\
M-A & Utt Mix A& &
background blend ratio & [0, 0.6] & linear & &
max background shift & [0, 30] & linear
\\
M-B & Utt Mix B& &
background blend ratio & [0, 0.6] & linear & &
background multiplicity & [0, 5] & linear
\\
\bottomrule
\end{tabular}
}
\label{t:speechstew}
\end{table*}

\subsection{Augmentation Operations}
\label{sec:aug_list}

We utilize 17 augmentation operations in our augmentation policies. We have listed all the augmentations and the significance of their two strength parameters $x_i$ in table \ref{t:bblocks}.

The identity operation is self-explanatory. Frequency masking, time masking and the variants thereof have been discussed in \cite{park2019specaugment,park2020specaugment}. We introduce two variants of time-warping \cite{park2019specaugment}, where the warping distance parameter is either absolute or determined with respect to the length of the utterance. Time perturbation uniformly stretches or shrinks the spectrogram in the time direction, effectively perturbing the speed of speech. Frequency warping is a warping deformation applied in the direction of the frequency axis. Cut out augmentation \cite{devries2017improved} applies multiple random rectangle masks to the spectrogram, whose the size and density are controlled by $x_i$. Frequency shift randomly shifts spectrogram values of frequency bands up or down, with the multiplicity and total coverage of shifted frequency bands controlled by $x_i$. Utterance mixing linearly mixes other utterances in the batch as background, with background blend ratio of $r_b$. Utterance mixing supports mixing multiple different backgrounds, as well as randomly shifting background utterances in the time dimension to introduce further stochasticity. Frequency noise samples Gaussian noise with a mean of 1.0 and stddev controlled by $x_1$ for each frequency band, and multiplies the noise scale vector to the features. Random convolution convolves input features with a random 2D depth-wise convolution filter that deviates from the identity filter by a random normal distribution with a standard deviation of 0.1. The frequency and time size of the filters are controlled by $x_i$.

\section{Methods}
\label{sec:experimental_setup}

In this section, we detail the set-up for G-Augment search experiments and baseline SpecAugment search experiments conducted on the warm-started CHiME-6 \cite{watanabe2020chime} and AMI \cite{ami} tasks. We then detail the set-up for exploring the transfer properties of the policies discovered by the search.

All experiments are conducted with ASR models utilizing a Conformer \cite{conformer} encoder. The encoder outputs are fed to RNN transducer \cite{graves2012sequence} with a 2-layer LSTM decoder. We use two models with respective model sizes of 100M and 1B parameters. We adapt the convention of denoting the two models as ConformerL and ConformerXXL following \cite{bigssl}.

\subsection{Search Experiments}

The augmentation policies are optimized on a ``search task" that can be undertaken with a manageable amount of resources, since training and evaluation on the task must be conducted many times. We find warm-started training of the ConformerL model on small corpora (CHiME-6 and AMI) adequate for this purpose.

\subsubsection{Pre-training}

The ConformerL model is prepared for the warm-start task by training on the SpeechStew dataset \cite{speechstew}. The SpeechStew dataset consists of seven public speech corpora---AMI \cite{ami}, Common Voice \cite{cv}, English Broadcast News%
\footnote{Linguistic data consortium (LDC) datasets LDC97S44, LDC97T22, LDC98S71 and LDC98T28.},
LibriSpeech \cite{librispeech}, Switchboard/Fisher%
\footnote{LDC datasets LDC2004T19, LDC2005T19, LDC2004S13, LDC2005S13 and LDC97S62.},
TED-LIUM v3 \cite{rousseau2012ted,hernandez2018ted} and Wall Street Journal\footnote{LDC datasets LDC93S6B and LDC94S13B.}. The utterances from these datasets are mixed randomly and batched for training. We discard the AMI portion of the dataset when preparing the model for the warm-start task on AMI. The model is trained by 100k steps with a batch size of 8192 on these datasets. Further details on the training parameters can be found in section 3 of \cite{speechstew}.

\subsubsection{Search Task: Warm-start Training}
\label{sec:warm_task}

We use two warm-start tasks for policy search, where we fine-tune the pre-trained ConformerL models of the previous section on the CHiME-6 \cite{watanabe2020chime}/AMI \cite{ami} datasets.

Given an augmentation policy $A$ we wish to evaluate, we fine-tune the pre-trained models with input augmented by $A$ for 30k steps via Adam optimization with $\beta_1=0.9$, $\beta_2=0.98$. We use a cosine learning rate schedule with 5k warm up steps and 2.5e-5 peak learning rate. This schedule reduces the variance of the final performance of the model, as the learning rate approaches zero by the end of training time. We train with a batch size of 256. We use the normalized word error rate (WER) of the development set computed at the final step of training to judge the fitness of the policy $A$.

For the CHiME-6 task, a global dropout rate of 0.5 is applied to all residual layers (attention, feed-forward, convolution), as well as feed-forward layers. For AMI, we use a global drop-out rate of 0.1. AMI comes with two dev/test sets---the ihm and sdm1 sets. We use the dev WER over both sets to evaluate the fitness.

\subsubsection{G-Augment Policy Search}

The G-Augment policy search is conducted using Vizier \cite{golovin2017google}, a distributed black-box optimization tool. The implementation of \cite{jaderberg2017population} of the generational evolutionary algorithm explained in section \ref{sec:g-augment:search_method} is utilized to optimize the policy. We use a population size $P=32$, a mutation rate of $\mu = 0.8$, and employ 64 parallel workers for 2000 trials. The search space is defined to be over augmentation graphs with $N = 25$ ensemble nodes where any of the 17 augmentations of section \ref{sec:aug_list} can be used. Each trial consists of training the warm-start task \ref{sec:warm_task} with an augmentation policy suggested by the algorithm, and judging the fitness by the dev-set WER of the trained network. Each trial requiring $\sim$ 10 hours to complete, a search experiment expends a total of 20k hours when run on 8 Google Cloud TPU chips, amounting to 160k TPU hours.

\subsubsection{Baseline: SpecAugment Random Search}

As a baseline, we conduct a random search over adaptive SpecAugment parameters with the equivalent compute budget of 160k TPU hours. This policy is obtained by composing a frequency mask and a time mask augmentation with adaptive multiplicity (see table \ref{t:bblocks}). The random search over the four strength parameters of these two augmentations is conducted across 2000 trials. 

\subsection{Policy Transfer Experiments}

Here we define some tasks for which transfer properties of an augmentation policy can be investigated.

\subsubsection{Cold-start Training}

We train ConformerL models on CHiME-6 and AMI from scratch without pre-trained initialization. Cold-start training trains with batch size 4096, using a transformer schedule with 10k warm-up steps and peak learning rate 2e-3. We use global dropout rate of 0.5/0.1 for CHiME-6/AMI respectively.

\subsubsection{Larger Models}

We also study transfer of the CHiME-6 augmentation policy to pre-trained ConformerXXL models. We have two set-ups for preparing the models, which have 10x parameters compared to the ConformerL. The first is the public set-up where the model is trained using wav2vec 2.0 \cite{baevski2020wav2vec} pre-training on the LibriLight dataset \cite{librilight} then further trained with the SpeechStew dataset. The other is the BigSSL \cite{bigssl} set-up where the model is trained with wav2vec 2.0 using an unlabeled YouTube dataset and further fine-tuned on a pseudo-labeled \cite{park2020nst1,xie2020nst2} YouTube dataset.

\subsubsection{Magnitude Tuning for Larger Models}
\label{sec:methods:mag_tuning}
Following \cite{cubuk2019randaugment}, we experiment with further tuning the overall magnitude of a policy when adapting it for training larger models. We use magnitude scaling, where we scale the magnitudes of all augmentations in the policy by a factor and round to the nearest integer. We consider scaling factors in the range of $[0.6, 1.4]$ in increments of $0.1$. We also consider magnitude increment, where we increment all magnitudes by an integer value within in $[-4, 4]$. The magnitudes are clipped within the range [0, 10].

\section{RESULTS}
\label{sec:majhead}

\subsection{Search Task Performance}
\label{sec:results:search_task}

We find that given the same computational budget, the G-Augment policies perform better than random-searched SpecAugment policies both for the CHiME-6 and AMI search tasks, as presented in tables \ref{t:chime_100M_results} and \ref{t:ami_100M_results}. This comparison is especially stark for CHiME-6, where the random search is not able to find a overall-beneficial SpecAugment policy.

\subsection{Policy Transfer to Cold-Start Training}
\label{sec:results:transfer_from_scratch}

The results for the transfer properties of the augmentation policies to cold-start training are presented in tables \ref{t:chime_100M_results} and \ref{t:ami_100M_results}. We find that the G-Augment policies transfer well to the cold-start training task. Also, the G-Augment policies show better performance against the random-searched SpecAugment baseline even on the transfer task. The transfer compares especially well for CHiME-6, where the random-searched SpecAugment policy turns out to be especially detrimental when applied to the cold-start task.

\begin{table}[h!]
  \caption{CHiME-6 WERs (\%) of ConformerL warm and cold-started models. The augmentation policies compared were obtained by searching on the warm-start task.}
  \label{t:chime_100M_results}
  \vskip 0.1in
  \label{t:c6ft}
  \centering
  \small
  \resizebox{0.85\columnwidth}{!}{%
  \begin{tabular}{lccccc}
    \toprule
    \bfseries Method &
    \multicolumn{2}{c}{Warm-start} & &
    \multicolumn{2}{c}{Cold-start} \\
    \cmidrule{2-3} \cmidrule{5-6}
    & dev & eval & & dev & eval \\
    \midrule
    \bfseries No Augmentation &  32.2 & 39.4 && 63.3 & 64.2 \\
    \midrule
    \bfseries With Augmentation \\
    \quad Manually Tuned SpecAugment \cite{speechstew} & 33.1 & 40.6 && 70.0 & 66.7 \\
    \quad Random-Searched SpecAugment & 32.0 & 40.2 && 69.9 & 65.1 \\
    \quad G-Augment Policy & \bfseries 31.7 & \bfseries 38.5 && \bfseries 62.0 & \bfseries 61.6 \\
    \bottomrule
  \end{tabular}
  }
\end{table}

\begin{table}[h!]
  \caption{AMI eval set WERs (\%) of ConformerL warm and cold-started models. The augmentation policies compared were obtained by searching on the warm-start task.}
  \label{t:ami_100M_results}
  \vskip 0.1in
  \label{t:c6ft}
  \centering
  \small
  \resizebox{0.85\columnwidth}{!}{%
  \begin{tabular}{lccccc}
    \toprule
    \bfseries Method &
    \multicolumn{2}{c}{Warm-start} & &
    \multicolumn{2}{c}{Cold-start} \\
    \cmidrule{2-3} \cmidrule{5-6}
    & ihm & sdm1 & & ihm & sdm1 \\
    \midrule
    \bfseries No Augmentation & 10.5 & 25.3 && 27.3 & 43.7 \\
    \midrule
    \bfseries With Augmentation \\
    \quad Random-Searched SpecAugment & 9.7 & 23.4 && 23.7 & 40.5 \\
    \quad G-Augment Policy & \bfseries 9.3 & \bfseries 23.3 && \bfseries 23.3 & \bfseries 40.1 \\
    \bottomrule
  \end{tabular}
  }
\end{table}

\subsection{Policy Transfer to Larger Models}

We next evaluate how well the G-Augment and random-searched SpecAugment policies transfer to larger models on the CHiME-6 task. We observe that no augmentation training provides a very strong baseline for this task.

\begin{table}[h!]
  \caption{CHiME-6 WERs (\%) from fine-tuning the ConformerXXL pre-trained with public data.}
  \label{t:chime_public}
  \vskip 0.1in
  \label{t:c6ft}
  \centering
  \small
  \resizebox{0.6\columnwidth}{!}{%
  \begin{tabular}{lccc}
    \toprule
    \bfseries Method & dev & eval \\
    \midrule
    \bfseries No Augmentation & \bfseries 30.7 & 39.3 \\
    \midrule
    \bfseries With Augmentation \\[2pt]
    \quad Random-Searched SpecAugment & 32.9 & 39.6 \\[2pt]
    \quad G-Augment Policy & 31.3 & 38.6 \\
    \qquad + Magnitude tuning & 30.9 & \bfseries 38.2 \\[2pt]
    \bottomrule
  \end{tabular}
  }
\end{table}

In table \ref{t:chime_public}, we present the results for policy transfer experiments to the public set-up. While the G-Augment policy compares favorably against the random-searched SpecAugment policy, we find that the performance is overall neutral. Upon applying a magnitude scaling factor of $0.8$, we are able to demonstrate improvement on the eval set.

We find that magnitude tuning is also necessary in order to gain meaningful improvement for the YouTube pre-trained set-up. We apply a magnitude increment of +1 to the G-Augument policy to achieve SoTA performance on CHiME-6 with 30.7$\%$ eval WER. Our results are summarized in table \ref{t:chime_sota_results}, where we compare our results against reported SoTA numbers in the literature. The last column of the table coincides with the status of overall SoTA performance on CHiME-6.

\begin{table}[h!]
  \caption{Comparing CHiME-6 WERs (\%) in three settings---ConformerL performance, ConformerXXL performance only utilizing additional public data and ConformerXXL performance without restrictions on the pre-training data.}
  \vskip 0.1in
  \label{t:chime_sota_results}
  \centering
  \small
  \resizebox{0.9\columnwidth}{!}{%
  \begin{tabular}{lcccccccc}
    \toprule
    &
    \multicolumn{2}{c}{L} & &
    \multicolumn{2}{c}{XXL Public} & &
    \multicolumn{2}{c}{XXL Any}\\
    \cmidrule{2-3} \cmidrule{5-6} \cmidrule{8-9}
    & dev & eval & & dev & eval & & dev & eval\\
    \midrule
    Previous SoTA \cite{speechstew,bigssl} & 33.1 & 40.6 && 31.9 & 38.9 &&
    26.2 & 31.0 \\
    G-Augment &
    \bfseries 31.7 & \bfseries 38.5 &&
    \bfseries 30.9 & \bfseries 38.2 &&
    \bfseries 26.0 & \bfseries 30.7 \\
    \bottomrule
  \end{tabular}
  }
\end{table}

\section{Discussion}

\noindent\textbf{Search Space Sizes:} The SpecAugment search space utilized has the size $11^4$ and 2000 random search trials cover close to 14\% of the search space. We find that the G-Augment search space considered is incomparably large to the SpecAugment search space used. For a crude estimate of a lower-bound of this search space, let us first assume that the application probability $q$ is quantized to have 11 values. Ignoring the graph structure search (assuming a linear graph structure), the selection probabilities (always assuming left edge selection), the second strength argument, the identity augmentation and even the order of augmentations while also imposing that all the augmentation parameter used should be distinct, we are still left with a search space size of ${M \choose N-1} \approx 10^{55}$ where $N-1=24$ is the number of left edges and $M = 16 * 11 * 11$ is the number of triples $(t, q, x_1)$.

\medskip

\noindent\textbf{Reducing the Search Cost:} In this work, we used an out-of-the-box search algorithm to demonstrate feasibility of our method. It would be interesting to apply cost-reduction methods for evolutionary search for G-Augment such as population based learning \cite{ho2019population}, functional equivalence checking or hurdles \cite{real2020automl} to reduce the computational cost.

\medskip

\noindent\textbf{CHiME-6:} We find that SpecAugment is surprisingly ineffective for CHiME-6, and discovering a good SpecAugment policy for this task is challenging.

\medskip

\begin{figure}[h!]
\centering
\includegraphics[width=0.8\columnwidth]{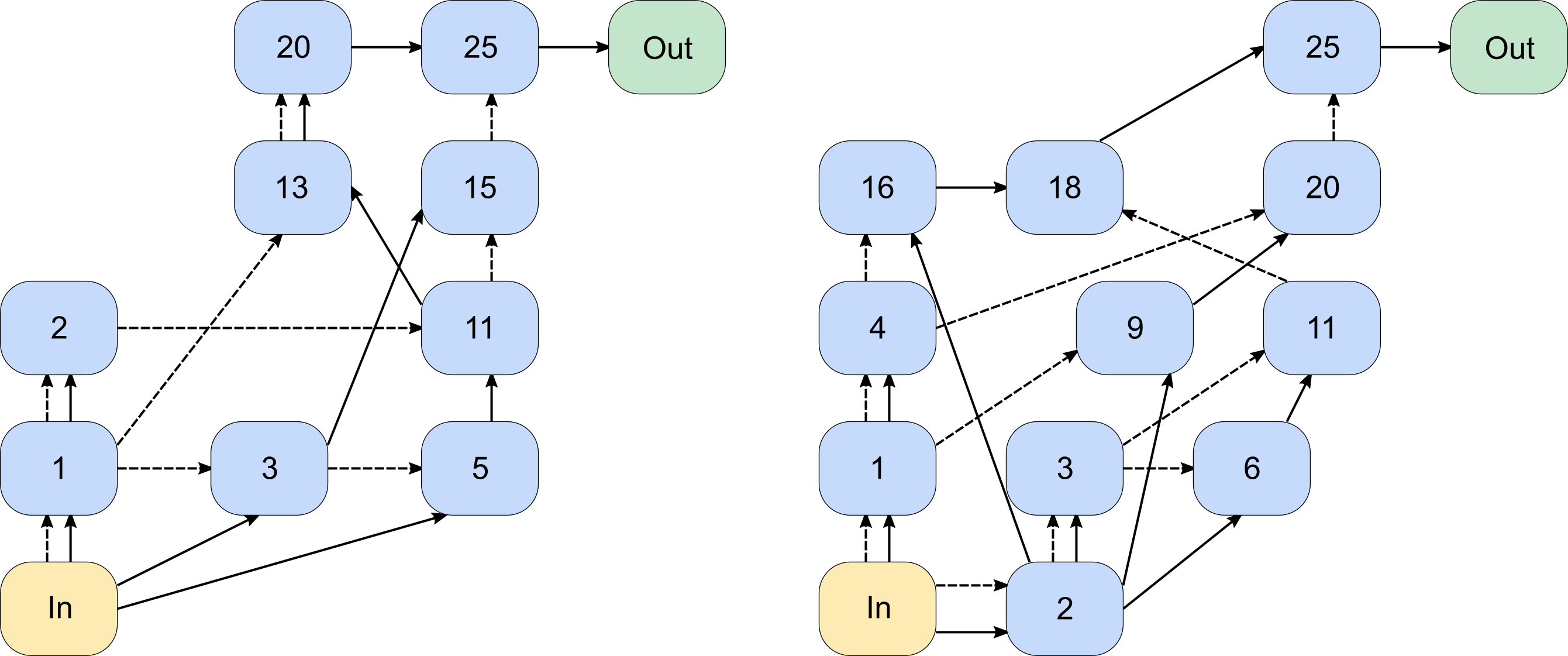}
\caption{Augmentation policy graphs discovered for CHiME-6 (left) and AMI (right). The dotted lines denote the left connections, while solid lines denote the right connections.}
\label{f:chime_ami}
\end{figure}

\noindent\textbf{Policy Analysis:} Here we present the policies discovered for CHiME-6 and AMI. The graphs obtained are plotted in figure \ref{f:chime_ami} where the dotted/solid lines have been used to indicate the left/right connections, respectively. The parameters associated to each edge have been listed in tables \ref{t:chime_policy} and \ref{t:ami_policy}.

\begin{table}[h!]
  \vskip -0.2in
  \caption{CHiME-6 Policy parameters.}
  \vskip 0.05in
  \label{t:chime_policy}
  \centering
  \small
  \resizebox{0.6\columnwidth}{!}{%
  \begin{tabular}{cccccccccc}
    \toprule
    Node & 1 & 2 & 3 & 5 & 11 & 13 & 15 & 20 & 25 \\
    \midrule
    $p_l$ & 0.5 & 0.5 & 0.0 & 0.5 & 0.5 & 0.9 & 0.8 & 0.4 & 0.5 \\
    $t_l$ & Id & TP & TM-AM & TM-AS & FW & RC & M-B & M-B & M-A \\
    $q_l$ & 1.0 & 0.39 & 0.25 & 0.45 & 0.65 & 0.14 & 0.53 & 0.96 & 0.67 \\
    $x_{1,l}$ & 4 & 7 & 9 & 5 & 4 & 6 & 4 & 2 & 9 \\
    $x_{2,l}$ & 4 & 6 & 2 & 7 & 10 & 8 & 3 & 4 & 6 \\
    \midrule
    $p_r$ & 0.5 & 0.5 & 1.0 & 0.5 & 0.5 & 0.1 & 0.2 & 0.6 & 0.5 \\
    $t_r$ & M-A & M-A & Id & FN & TP & FN & Id & FS & TP \\
    $q_r$ & 0.17 & 0.004 & 0.47 & 0.75 & 0.05 & 0.28 & 0.87 & 0.67 & 0.24 \\
    $x_{1,r}$ & 2 & 4 & 2 & 9 & 5 & 8 & 9 & 7 & 3 \\
    $x_{2,r}$ & 1 & 4 & 0 & 8 & 0 & 9 & 9 & 4 & 9 \\
    \bottomrule
  \end{tabular}
  }
\end{table}

We can observe that only a single time-masking augmentation is used in the CHiME-6 policy and no frequency maskings are utilized. This should be contrasted with the AMI policy where time and frequency maskings are amply utilized.

\begin{table}[h!]
  \caption{AMI Policy parameters.}
  \vskip 0.05in
  \label{t:ami_policy}
  \centering
  \small
  \resizebox{0.9\columnwidth}{!}{%
  \begin{tabular}{cccccccccccc}
    \toprule
    Node & 1 & 2 & 3 & 4 & 6 & 9 & 11 & 16 & 18 & 20 & 25 \\
    \midrule
    $p_l$ & 0.9 & 0.1 & 0.1 & 0.3 & 0.9 & 0.3 & 0.3 & 0.1 & 0.0 & 0.2 & 0.8 \\
    $t_l$ & TM-AM & FM & TM-AM & FS & TM-AM & TM-AS & TM-FA & TM-FA & TM-AS & TM-AM & TM-AS \\
    $q_l$ & 1.00 & 0.77 & 0.46 & 0.19 & 0.17 & 0.36 & 0.76 & 0.99 & 0.56 & 0.64 & 0.30 \\
    $x_{1,l}$ & 8 & 1 & 9 & 4 & 9 & 5 & 6 & 7 & 4 & 8 & 5 \\
    $x_{2,l}$ & 7 & 10 & 5 & 4 & 9 & 5 & 3 & 1 & 3 & 4 & 9 \\
    \midrule
    $p_r$ & 0.1 & 0.9 & 0.9 & 0.7 & 0.1 & 0.7 & 0.7 & 0.9 & 1.0 & 0.8 & 0.2 \\
    $t_r$ & CO & M-B & TW-A & Id & FN & FW & TW & FW & M-A & TM-AS & FS \\
    $q_r$ & 0.31 & 0.74 & 0.52 & 0.49 & 0.92 & 0.61 & 0.04 & 0.42 & 0.64 & 0.40 & 0.10 \\
    $x_{1,r}$ & 5 & 9 & 5 & 9 & 1 & 0 & 5 & 8 & 9 & 6 & 4 \\
    $x_{2,r}$ & 3 & 10 & 0 & 8 & 2 & 5 & 9 & 6 & 2 & 1 & 7 \\
    \bottomrule
  \end{tabular}
  }
\end{table}

\section{Conclusion}

We have presented G-Augment, a framework to simultaneously optimize the structure of a data augmentation policies along with its parameters with an evolutionary algorithm. G-Augment structures augmentation policies as a directed acyclic graph, enabling a definition and discovery of stochastic policies with a hierarchical structure. While we have demonstrated the utility of G-Augment for ASR tasks, it may be employed for any task in which the optimization of augmentation policies can lead to improvement in performance.

\section{ACKNOWLEDGMENTS}
\label{sec:ack}

We would like to thank Zhehuai Chen, Chung-Cheng Chiu, Mike Chrzanowski, Yu Zhang and Barret Zoph for collaboration on earlier iterations of this work. We thank William Chan, Xuanyi Dong and Esteban Real for helpful discussions.


\bibliographystyle{IEEEbib}
\small
\bibliography{mybib}

\end{document}